\pdfoutput=1

\documentclass[11pt]{article}

\usepackage[final]{acl}

\usepackage{times}
\usepackage{latexsym}

\usepackage[T1]{fontenc}

\usepackage[utf8]{inputenc}

\usepackage{microtype}

\usepackage{inconsolata}
\usepackage{xspace}

\usepackage{graphicx}
\usepackage{amssymb}
\usepackage{amsmath}
\usepackage{amssymb}
\usepackage{mathtools}
\usepackage{amsthm}
\usepackage{url}
\usepackage{booktabs}

\title{Multi-Objective Linguistic Control of Large Language Models
}

\author{Dang Nguyen, Jiuhai Chen, Tianyi Zhou\\
    University of Maryland, College Park\\
    \texttt{\{dangmn, jchen169, tianyi\}@umd.edu} \\
    Project: \url{https://github.com/tianyi-lab/mctune}
}

\begin{document}
\maketitle
\begin{abstract}
Large language models (LLMs), despite their breakthroughs on many challenging benchmark tasks, lean to generate verbose responses and lack the controllability of output complexity, which is usually preferred by human users in practice. In this paper, we study how to precisely control multiple linguistic complexities of LLM output by finetuning using off-the-shelf data. To this end, we propose multi-control tuning (MCTune), which includes multiple linguistic complexity values of ground-truth responses as controls in the input for instruction tuning. We finetune LLaMA2-7B on Alpaca-GPT4 and WizardLM datasets. Evaluations on widely used benchmarks demonstrate that our method does not only improve LLMs' multi-complexity controllability substantially but also retains or even enhances the quality of the responses as a side benefit. \looseness-1
\end{abstract}

\section{Introduction}
Large language models have achieved remarkable success in generating free-from texts for different downstream tasks or human instructions \cite{brown2020language, chowdhery2022palm, touvron2023llama}. However, existing LLMs still lack precise control over the linguistic complexity of their outputs \cite{DBLP:journals/corr/abs-2305-18449, chen2024benchmarking, DBLP:journals/corr/abs-2402-10614}, e.g., the total number of nouns, the variation of verbs, etc. 
Linguistic controllability is crucial to creating personalized outputs since those complexity indices directly reflect human reading complexity in multiple aspects. For example, a short yes/no answer is required by some users while a detailed explanation is preferred by others. 
Moreover, recent studies \cite{DBLP:journals/corr/abs-2310-03716, DBLP:journals/corr/abs-2402-04833, DBLP:journals/corr/abs-2403-13787} have discovered a spurious correlation between the quality reward used in LLM alignment and the output length (i.e., a specific linguistic complexity). 
Consequently, LLMs favor generating verbose responses due to the length bias, which may increase unnecessary reading complexity. 
It is still an open problem to mitigate the bias without hurting the output quality.

While existing LLM finetuning techniques such as instruction-tuning \cite{brown2020language, DBLP:conf/nips/StiennonO0ZLVRA20, DBLP:journals/corr/abs-2402-10110} and reinforcement learning from human feedback (RLHF) \cite{DBLP:conf/nips/ChristianoLBMLA17, DBLP:journals/corr/abs-2204-05862} have been demonstrated to be effective in aligning the output with human intent or preference, they only focus on maximizing a single objective. 
Instead, achieving the controllability of multiple complexity indices requires a non-trivial multi-objective optimization that has not been thoroughly studied on LLMs. 
Rather than solely maximizing or minimizing the complexities, it aims to reach different target complexity values on the Pareto frontier. 
This requires LLMs to adjust the trade-off among objectives and capture their potential correlations or constraints in the text generation process. 
In addition, due to the huge space of possible combinations of complexity indices, it could be expensive to collect training data for multi-objective control. 

In this paper, we take the first step towards multi-objective control of the linguistic complexity of LLM outputs. Instead of collecting new data, our strategy allows the reuse of existing instruction-tuning data. In particular, we annotate the ground-truth responses in a dataset by their complexity metrics evaluated using tools developed in computational linguistics. The multiple complexity indices and their values are appended as tags to the input so finetuning an LLM on the linguistic-label augmented data helps build a strong connection between the input tags and the linguistic complexity of the output, hence enforcing the LLM to adhere to the complexity requirements during sequential decoding. Surprisingly, we observe that randomly sampling a small subset of tags for each training example suffice to obtain controllability over all the complexities of test examples, thereby reducing the required amount of training data. 

We examine our approach by finetuning LLaMA2-7B \cite{DBLP:journals/corr/abs-2307-09288} using a linguistic-complexity labeled Alpaca-GPT4 dataset \cite{peng2023instruction} (i.e., the prompts are from the original Alpaca dataset while the responses are composed by GPT-4) and a WizardLM dataset \cite{xu2023wizardlm}. We do not only observe an expected substantial improvement in linguistic controllability but also a side benefit of enhanced response quality, indicating that finetuning under multiple linguistic constraints can improve the LLM for general purposes. Compared to unconstrained RLHF and IFT methods, which suffer from the length bias \citep{DBLP:journals/corr/abs-2402-07319}, our approach does not introduce the bias but can improve both the controllability and quality simultaneously. 

\section{Related Work}

\paragraph{Linguistic Features and Complexity.}
The exploration of linguistic features and complexity in language models encompasses a diverse range of research. Seminal studies have investigated the syntactic abilities of LSTMs \citep{linzen2016assessing}. \citet{ficler-goldberg-2017-controlling} has introduced methods for manipulating the stylistics and syntactic output of test generation models. Additionally, linguistic style transfer \citep{shen2017style} has also showcased the  adaptability of language models to capture and replicate varied linguistic features. Building on the previous efforts to understand the multifaceted nature of linguistics complexity, our work concentrates on producing responses endowed with specific linguistic characteristics, which is less explored in the previous work.

\paragraph{Controllability of LLMs.} 
The topic of personalized language modeling has attracted significant attention across various research papers.
Techniques such as user embedding have become common for customizing language models to individual needs \cite{welch-etal-2020-exploring, rocca-yarkoni-2022-language}. More recently, \citet{mireshghallah-etal-2022-useridentifier, oba2023perplm} propose prompt based personalized fine-tuning for specific users, and producing personalized responses. Our research shifts the focus from personalization for specific users to the broader goal of controlling large language models (LLMs) to produce outputs with  linguistic diversity and complexity, addressing a gap not explored by the aforementioned works.

Another prevalent technique for guiding the output of large language models involves Tagging \citep{pmlr-v202-korbak23a, prabhumoye2023adding, lu2022quark}. This method incorporates appending human-readable text during the training of LLMs. Contrary to previous studies that concentrated on managing aspects like toxicity \citep{pmlr-v202-korbak23a, prabhumoye2023adding, DBLP:journals/corr/abs-2402-10614} and controlling repetition \citep{ lu2022quark}, our approach employs tagging techniques to control linguistic features across multiple attributes. Additionally, we utilize multiple tags to enable simultaneous consideration of various attributes, a difference from earlier work that primarily uses of a single tag.

\paragraph{Finetuning and Alignment of LLMs.} 

With the emergence of large-scale language models, such as those in the GPT series, aligning language models has become prevalent.
Studies like those by \citep{zhou2023lima, xu2023wizardlm, Li2023FromQT, DBLP:journals/corr/abs-2402-00530, DBLP:journals/corr/abs-2403-12776} have concentrated on the process of data curation for instruction fine-tuning to enhance models' instructions following capabilities. Unlike these data-centric approaches, we keep the original instruction dataset \citep{DBLP:journals/corr/abs-2308-16175} but augment instructions with various tags to introduce a richer array of linguistic features, thereby elevating the instruction-following capabilities of the models.

\section{Finetuning LLMs for Linguistic Controllability}
In this section, we delineate our approach to multi-control tuning. Section \ref{lftk} outlines the linguistic features of interest and describes how we extract them from a given text segment. In Section \ref{mctune}, we explain how the extracted features are incorporated into the multi-control tuning process.

\subsection{Handcrafted Linguistic Features} \label{lftk}
\begin{table}[]
\centering
\resizebox{\columnwidth}{!}{%
\begin{tabular}{c|l|l}
\hline
ID & Name                 & Description                 \\ \hline
1  & \texttt{t\_word    } & number of words             \\
2  & \texttt{n\_noun    } & number of nouns             \\
3  & \texttt{n\_verb    } & number of verbs             \\
4  & \texttt{n\_adj     } & number of adjectives        \\
5  & \texttt{t\_uword   } & number of unique words      \\
6  & \texttt{n\_unoun   } & number of unique nouns      \\
7  & \texttt{n\_uverb   } & number of unique verbs      \\
8  & \texttt{n\_uadj    } & number of unique adjectives \\
9  & \texttt{ttr        } & type-token ratio            \\
10 & \texttt{noun\_var  } & noun variation              \\
11 & \texttt{verb\_var  } & verb variation              \\
12 & \texttt{adj\_var   } & adjective variation         \\
13 & \texttt{fkre       } & Flesch-Kincaid reading ease \\
14 & \texttt{rt\_average} & average reading time        \\ \hline
\end{tabular}%
}
\caption{The list of linguistic features for controllability tuning. The second column shows the name of each feature with the corresponding descriptions in the third column. We include a detailed explanation of how these features are computed in Appendix~\ref{sec:compute_features}.}
\label{tab:features}
\end{table}

We are specifically interested in controlling the \textit{handcrafted} linguistic properties of the model's generation. This type of feature has been used throughout the NLP field \cite{bogdanova-etal-2017-cant, choudhary2021linguistic, lee2021pushing} and is loosely defined in \citet{lee-lee-2023-lftk} as \textit{``a single numerical value produced by a uniquely identifiable method on any natural language.''} An example of a linguistic feature not considered handcrafted is text embeddings produced by deep neural networks, which usually take the form of a vector. We extract such features from a text segment using the LFTK package proposed in \citet{lee-lee-2023-lftk}. It encompasses a diverse set of 220 features that are grouped into different linguistic families. Within the scope of this paper, we sample a reasonably-sized set of 14 features for multi-control tuning, which are presented in Table \ref{tab:features}. These features are selected to cover most of the feature families while being simple to understand and verify by human users.

Formally, given a text segment $\mathbf{x} = [x_1, x_2, ..., x_l]$, with $x_i$ being the $i$-th token of $\mathbf{x}$, we denote the feature extractor as a function $f: \mathcal{X} \rightarrow \mathbb{R}^d$ that maps $\mathbf{x}$ to a $d$-dimensional vector $f(\mathbf{x}) = \left[f_1(\mathbf{x}), ..., f_d(\mathbf{x})\right]$, where $f_j(\mathbf{x})$ represents the $j$-th linguistic feature of interest and $\mathcal{X}$ represents the space of all texts. In this paper, the function $f$ refers to the LFTK feature extractor.

\subsection{Multi-Objective Control Tuning} \label{mctune}
Consider the standard instruction tuning setting where an LLM, denoted as $p_\theta(\mathbf{x}) = \prod^{l}_{i=1} p_\theta(x_i\mid x_{<i})$, is trained on a dataset of $N$ instruction-output pairs, $D_\text{train}=\left\{(\mathbf{x}_i, \mathbf{y}_i)\right\}^{N}_{i=1}$. The training objective is to generate $\mathbf{y}_i$ given $\mathbf{x}_i$; thus, the loss is
$$\mathcal{L} = - \sum^{N}_{i=1}\sum ^{|\mathbf{y}_i|}_{k=1}\log{p_\theta(\mathbf{y}_{i, k} \mid \mathbf{x}_i, \mathbf{y}_{i, <k})}$$
We address multi-control tuning by framing it as a conditional instruction tuning problem, where we utilize the target response $\mathbf{y}_i$ from $D_\text{train}$ to generate a linguistic control vector $f(\mathbf{y}_i)$ and append it to $\mathbf{x}_i$. The model is then trained to generate $\mathbf{y}_i$ condition on both $\mathbf{x}_i$ and $f(\mathbf{y}_i)$. However, to enhance data diversity and better simulate real-world scenarios, where a user may wish to control only a few features, we do not utilize all features for every data example. For each pair $(\mathbf{x}_i, \mathbf{y}_i) \in D_\text{train}$, we randomly sample an integer $n_i \sim \operatorname{Uniform}\left\{1, \ldots, m\right\},\, m \leq d$. The set of $n_i$ feature indices, denoted by $C_i$, is then randomly sampled from the pool of all feature indices, i.e., $C_i \sim \operatorname{Uniform}\left(\left\{A \subseteq \left\{1,\ldots,d\right\}: |A|=n_i\right\}\right)$. The linguistic control vector used for the $i$-th example is denoted as $f_{C_i}(\mathbf{y}_i)$, where $f_{C_i}(\mathbf{y}_i) = \left[ f_{C_{i,1}}(\mathbf{y}_i), \ldots, f_{C_{i,n_i}}(\mathbf{y}_i) \right]$. The resulting training loss becomes
$$\mathcal{L} = - \sum^{N}_{i=1}\sum ^{|\mathbf{y}_i|}_{k=1}\log{p_\theta(\mathbf{y}_{i, k} \mid \mathbf{x}_i, f_{C_i}(\mathbf{y}_i), \mathbf{y}_{i, <k})}$$
The benefits of our training strategy are twofold: (1) it improves the controllability and instruction-following capability simultaneously, thus avoiding the catastrophic forgetting problem that may degrade the model's generation quality. In fact, we will later show in Section \ref{sec:quality_eval} that LLMs trained with our approach achieve even stronger instruction-following ability compared to those trained with vanilla instruction tuning; (2) it allows us to utilize off-the-shelf datasets without the need to collect new ones.

\subsection{Prompt Template}
\begin{figure*}[h]
\begin{center}
\centerline{\includegraphics[width=\textwidth]{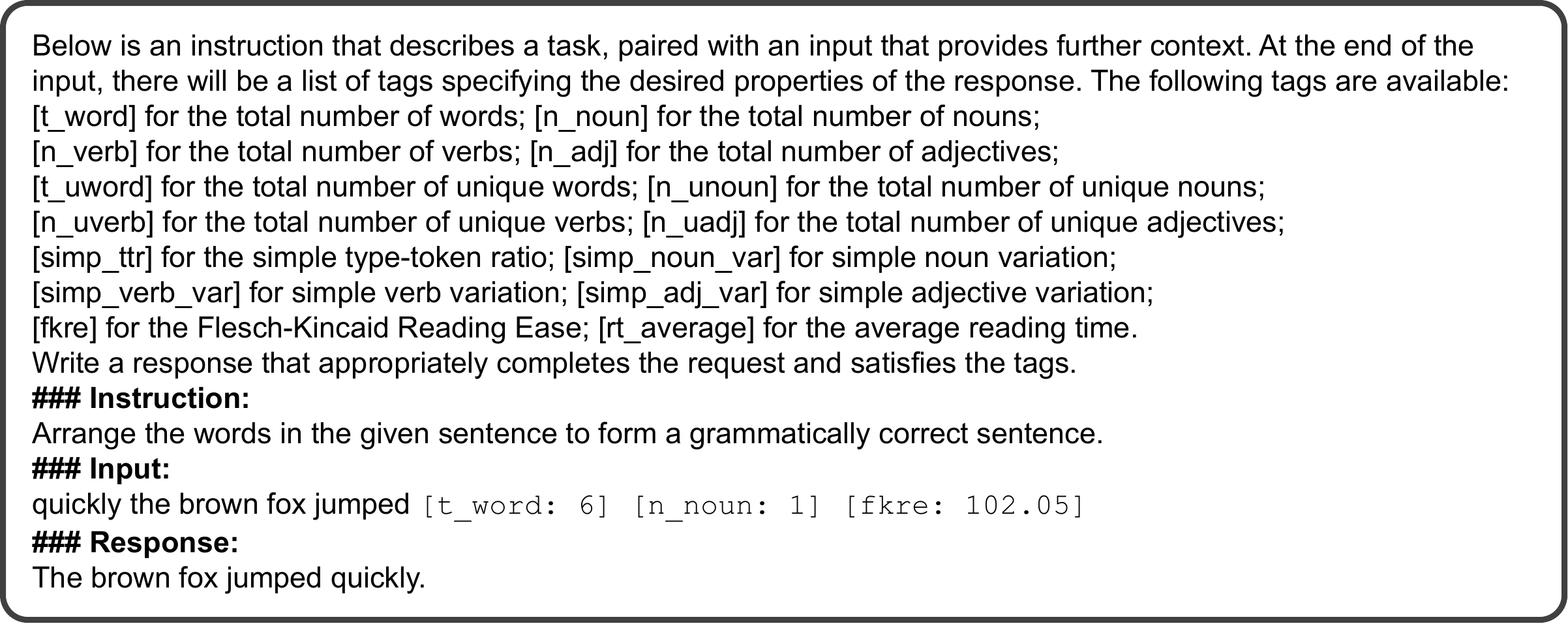}}
\caption{An example of how data is formatted before being fed into LLMs in this paper. The first paragraph presents a system prompt containing a complete list of feature descriptions. Our preliminary results indicate that including descriptions enhances the effectiveness of our approach.}
\label{fig:prompt_template}
\end{center}
\vspace{-1em}
\end{figure*}

In practice, we format both the instruction $\mathbf{x}$ and the output $\mathbf{y}$ into a predefined template before they are input into the LLMs. This paper adopts the template outlined in \cite{alpaca}, wherein $\mathbf{x}$ is decomposed into an instruction and an input component. Regarding linguistic controls, we format them into a sequence of the form $\texttt{[name\_1: value\_1] ... [name\_n: value\_n]}$, aiming for conciseness by utilizing the features' abbreviated names listed in Table \ref{tab:features}. To assist the LLM in better understanding these abbreviations, we provide a comprehensive list of feature descriptions within the system prompt, which has been confirmed to enhance the effectiveness of our approach in preliminary experiments. This sequence of controls is subsequently attached to the input component. Figure \ref{fig:prompt_template} illustrates a detailed example of how an input prompt is constructed.

\section{Evaluation of Linguistic Controllability} \label{control_eval}
During evaluation on a reserved test dataset $D_\text{test}$, we aim to measure how the model performs as the linguistic control vector $f$ changes. To conduct such an evaluation, we develop a sampling strategy to sample different control vectors for an instruction $\mathbf{x}_i$. Specifically, we need a sampling strategy that: (1) is specific to an instruction $\mathbf{x}_i$, i.e., the sampled control vectors should not stray too far from the reasonable range for a specific $\mathbf{x}_i$. For example, an instruction to "Generate a short story" should not have a large value for \texttt{t\_word}; (2) ensures the sampled control vectors are always valid, meaning that no linguistic controls conflict with each other (e.g., \texttt{t\_uword} should always be less than or equal to \texttt{t\_word}), and no control is out-of-bound (e.g., \texttt{fkre} should always be less than or equal to 121.22).

To achieve the first goal, we utilize $\mathbf{y}_i$'s linguistic feature vector as a reference point and sample new control vectors $f'$ from the Gaussian distribution centered at $f(\mathbf{y}_i)$, i.e., $f' \sim \mathcal{N}(f(\mathbf{y}_i), I\sigma^2)$. However, since each feature $f_i(\mathbf{y})$ has a different range, a small $\sigma$ for some features may be large for others. To avoid this inconsistency, we standardize all features to unit variance before sampling. More formally, the new control vector $f'$ is computed by
$$f' = z^{-1}(z(f(\mathbf{y}_i)) + \sigma\epsilon),\,\,\,\,\epsilon \sim \mathcal{N}(0, I)$$
where $z: \mathbb{R}^d \rightarrow \mathbb{R}^d$ standardizes each feature to unit variance, and $z^{-1}$ is the inverse operation.

While sampling a valid control vector can be a challenging problem, verifying its validity is straightforward. This can be done using a simple rule-based method, which we outline in Appendix \ref{sec:verify_validity}. We utilize this observation to achieve the second goal by performing rejection-based sampling, i.e., keep resampling a new control vector $f'$ until we find a valid one. We will show in Section \ref{sec:sigma_impact} that $\sigma$ can serve as a hyperparameter to control the evaluation difficulty.

Lastly, for each example in $D_\text{test}$, we randomly sample a number $n$, a set of $n$ feature indices $C$, and $K$ new control vectors $f'_C$ for controllability and generation quality evaluation.

\section{Experiments}
\subsection{Implementation Details}
By default, we set $K=5$ and $\sigma=0.1$ unless specified otherwise. We set the maximum number of linguistic controls per example to $m=5$ and will show in Section \ref{sec:ablation_num_tags} that limiting $m$ to 5 does not affect the model's controllability, even when more than 5 controls are used in the evaluation. For all datasets, we fine-tune LLaMA2-7B using the AdamW optimizer with a linear learning rate schedule. We set the learning rate to $2\times10^{-5}$, the batch size to 128, and the number of warmup steps to 100. All models are trained on 8 RTX A6000 GPUs for 5 epochs.

\subsection{Datasets.}
\paragraph{Data Preprocessing.} Because LFTK cannot extract linguistic features with perfect accuracy, it sometimes generates invalid control vectors \(f(\mathbf{y}_i)\) (e.g., producing an \texttt{fkre} greater than 121.22). This situation complicates the process of sampling new control vectors, as \(f(\mathbf{y}_i)\) may deviate significantly from the feasible region, drastically reducing the probability of sampling a valid one. Therefore, we exclude data examples with invalid \(f(\mathbf{y}_i)\).

\paragraph{Alpaca-GPT4.} We utilize the Alpaca-GPT4 dataset \cite{peng2023instruction} for instruction tuning and evaluation. This dataset shares the same set of instructions as the Alpaca dataset \cite{alpaca}, but it employs OpenAI's GPT-4 to generate high-quality responses. Our preliminary experiments demonstrate that training with this dataset yields better results in terms of both controllability and generation quality compared to training with the original Alpaca dataset. After preprocessing, the dataset is divided into a training set of 45,000 examples and a test set of 2,000 examples.

\paragraph{WizardLM.} In addition to Alpaca-GPT4, we also evaluate our method on the WizardLM dataset \cite{xu2023wizardlm}. The original dataset contains 70,000 examples of instruction-output pairs that were automatically generated by ChatGPT. We subsampled 50,000 examples from the original dataset. Similar to Alpaca-GPT4, after preprocessing, we split the data into a training set of 40,000 examples and a test set of 2,000 examples.

\subsection{Models.}
We use LLaMA2-7B \cite{touvron2023llama} as the base model for multi-control tuning in all experiments. 

\subsection{Baselines}
To evaluate the impact of our method on controllability and generation quality, we conduct comparisons with the same LLaMA2-7B base model but trained with regular instruction fine-tuning. We also include OpenAI's GPT-3.5 Turbo (gpt-3.5-turbo-0125) as a baseline to compare our model against state-of-the-art proprietary LLMs in terms of controllability. For each baseline, we control the linguistic complexity of their responses using the prompt template shown in Figure~\ref{fig:prompt_template} in a zero-shot manner.

\subsection{Evaluation Metrics} \label{eval_metrics}
\paragraph{Controllability Error.}
Given an instruction $\mathbf{x}$, a linguistic control vector $f_{C}$, and a generated response $\hat{\mathbf{y}} \sim p_\theta(\mathbf{y}\mid \mathbf{x}, f_C)$, we measure controllability error by computing the $L_1$ error between the specified control vector $f_{C}$ and the response's linguistic feature vector $f_{C}(\hat{\mathbf{y}})$. We denote this error as $\mathbf{e} = \left| f_{C}(\hat{\mathbf{y}}) - f_{C}\right| \in \mathbb{R}^{|C|}$.

\paragraph{Quality Score.}
To evaluate generation quality, we follow \citet{zheng2023judging} in using a powerful LLM (e.g., ChatGPT-4 Turbo) as a judge to assign quality scores ranging from 1 to 10.

\subsection{Main Results and Analysis}
\subsubsection{Linguistic Controllability Evaluation}
\begin{figure*}[h]
\begin{center}
\centerline{\includegraphics[width=\textwidth]{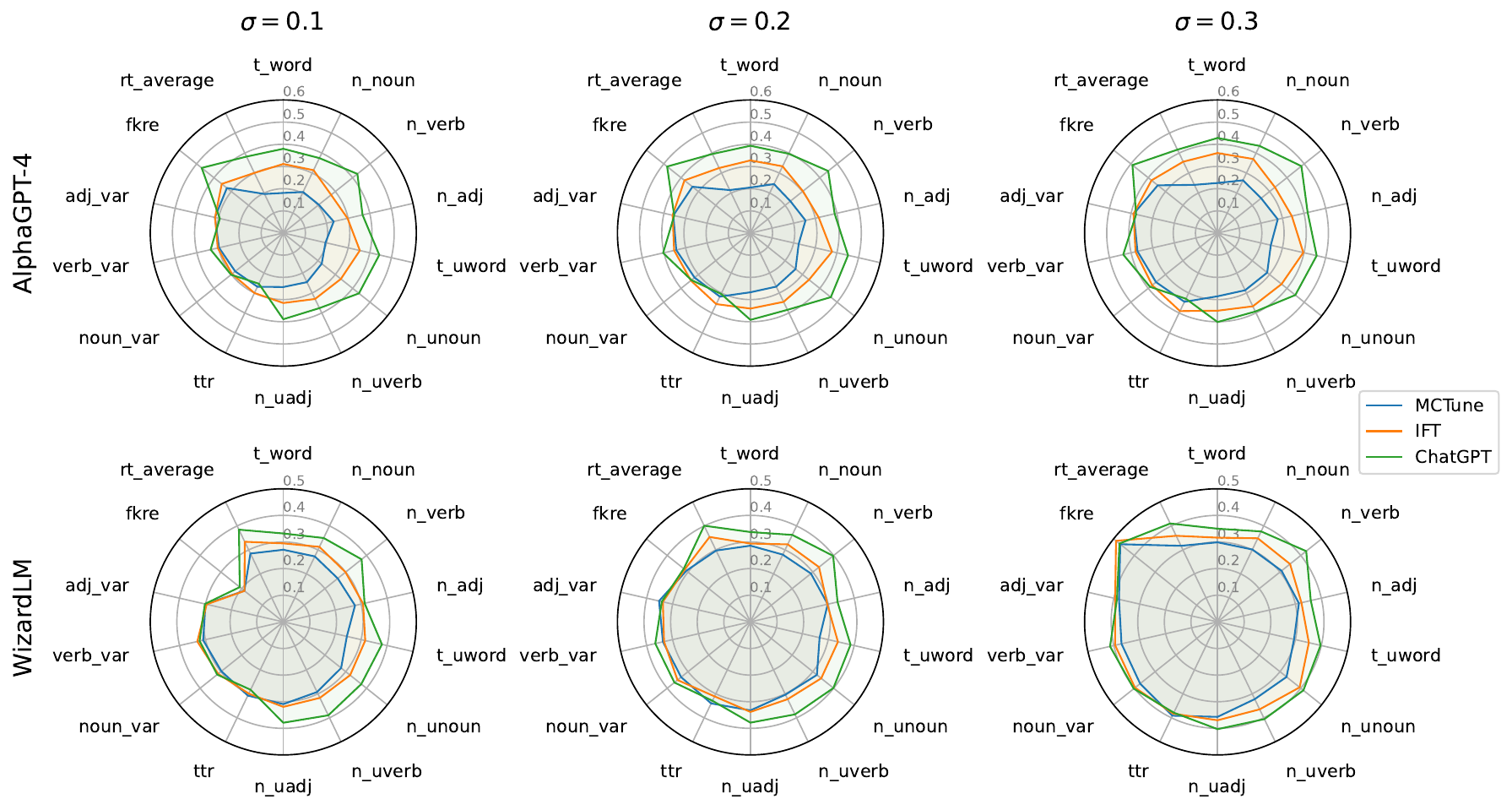}}
\caption{\textbf{Comparison of controllability error} (the average normalized \(L_1\) error) of ChatGPT (\texttt{gpt-3.5-turbo-0125}), IFT (finetuning without controls), and MCTune (ours) on AlpacaGPT-4 and WizardLM datasets and three test settings of target linguistic complexity with increasing \(\sigma\) (difficulty). To visualize each linguistic feature's average error on the same radar plot, we apply min-max normalization to normalize them to a similar scale. To reduce the effect of outliers, the minimum \(L_1\) error of the \(i\)-th feature is the minimum among all baselines, and the maximum \(L_1\) error refers to the 95th percentile of errors among all baselines.
}
\label{fig:main_results}
\end{center}
\vspace{-1em}
\end{figure*}
This section presents the evaluation of linguistic controllability between our method and various baselines. We consider three evaluation settings: Easy ($\sigma=0.1$), Medium ($\sigma=0.2$), and Hard ($\sigma=0.3$). Our goal is to measure the controllability error of each baseline on each linguistic control and then visualize all of them on the same radar plot for comparison. Formally, for each linguistic control $i$, and each baseline $j$, we maintain a list $\mathbf{e}_{i, j} = \left[e^{1}_{i, j}, \ldots, e^{|\mathbf{e}_{i, j}|}_{i, j}\right]$ of $L_1$ errors made by $j$, where the length of this list depends on the number of text examples that contain $i$. The matrix of all $L_1$ errors for feature $i$ is denoted as
\begin{equation}
\label{error}
\mathbf{E}_i = \begin{pmatrix}
e^{1}_{i, 1} & \ldots & e^{|\mathbf{e}_{i, 1}|}_{i, 1} \\
\vdots & \ddots  & \vdots  \\
e^{1}_{i, J} & \ldots & e^{|\mathbf{e}_{i, J}|}_{i, J} \\
\end{pmatrix}    
\end{equation}
It is challenging to directly visualize $\mathbf{E}_i$ for all $i$ on the same radar plot because each linguistic control $i$ has a different range of values. Therefore, we normalize each element in $\mathbf{E}_i$ using min-max normalization, with the minimum being $\min{E_i}$ and the maximum being the 95th percentile of $\mathbf{E}_i$, or $P_{95}(\mathbf{E}_i)$. For each baseline $j$, its average normalized $L_1$ error on control $i$ is calculated as $\sum_k \operatorname{norm}(e^k_{i, j}) / |\mathbf{e}_{i,j}|$, where $\operatorname{norm}(\cdot)$ denotes the normalization described above.

As shown in Figure \ref{fig:main_results}, our method consistently outperforms the other baselines across all settings. Surprisingly, a state-of-the-art model like ChatGPT underperforms in controllability compared to the instruction-finetuned LLaMA2-7B. Upon manual inspection, we observed that ChatGPT tends to produce more verbose responses regardless of the linguistic controls applied, which may explain the observed poor performance. Another possible reason for this discrepancy is that LLaMA2-7B is finetuned on data sharing the same distribution as the test set, giving it an advantage over ChatGPT. Note that for \texttt{noun\_var}, \texttt{verb\_var}, and \texttt{adj\_var}, the differences between each baseline's errors are not significant. We hypothesize that, in contrast to other linguistic complexities, the descriptions of these features are somewhat more ambiguous, and we also did not provide a clear description in the system prompt, which may not be helpful for the model to understand and follow these features.
\begin{figure*}[h]
\begin{center}
\centerline{\includegraphics[width=\textwidth]{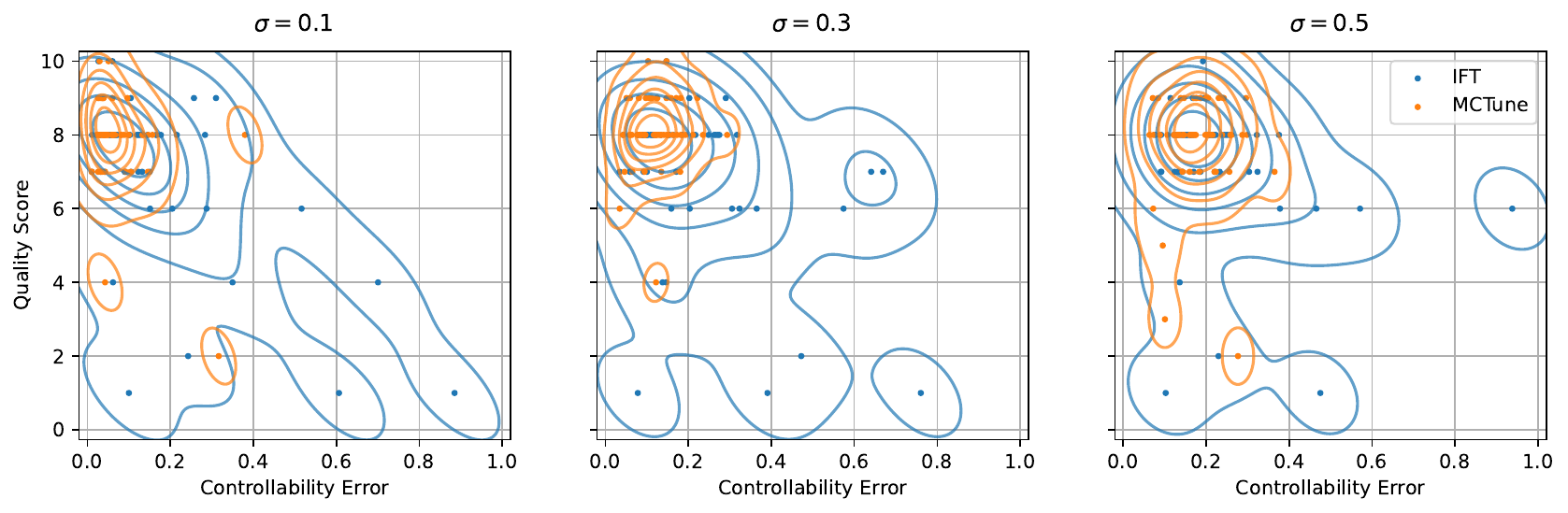}}
\caption{\textbf{Trade-off between linguistic controllability and generation quality} in three increasingly difficult test settings. Each dot represents a model's response $\hat{\mathbf{y}}$ to a specific query $\left[\mathbf{x}, f_C\right]$. The response is given a quality score from 1 to 10 by a judge LLM (GPT-4 Turbo) based on how well $\hat{\mathbf{y}}$ addresses $\mathbf{x}$. A controllability error is measured for $\hat{\mathbf{y}}$, which is computed by taking the average of normalized $L_1$ errors across all linguistic controls in $f_C$. Blue and orange dots respectively represent responses from models trained by IFT and MCTune using Alpaca-GPT4 dataset.\looseness-1}
\label{fig:quality_vs_controllability}
\end{center}
\end{figure*}

\subsubsection{Generation Quality Evaluation} \label{sec:quality_eval}
\begin{figure}[h]
\begin{center}
\centerline{\includegraphics[width=7cm]{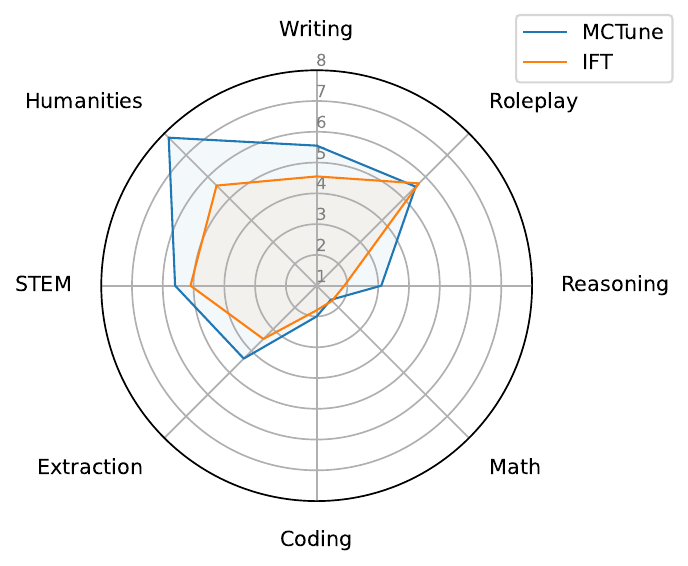}}
\caption{\textbf{Comparison between MCTune and IFT-trained models on MT-Bench.} We finetune LLaMA2-7B on Alpaca-GPT4 dataset and GPT-4 Turbo is the judge in the test. The average score per axis ranges from 1 to 10 and are given by the judge.}
\label{fig:quality_results}
\end{center}
\vspace{-1em}
\end{figure}

A natural concern when fine-tuning for controllability is how it affects the model's generation quality. In this section, we answer this question by comparing our method with standard instruction fine-tuning on MT-Bench \cite{zheng2023judging}, a widely used benchmark for evaluating LLMs on multi-turn open-ended questions. Specifically, we use GPT-4 Turbo (\texttt{gpt-4-0125-preview}) as a judge and compare the two methods in both single-answer and pairwise settings. In the single-answer setting, the judge assigns a single numeric score from 1 to 10 for each model's answer. In the pairwise setting, the judge receives two answers from both baselines and returns either a win, a loss, or a tie. The result for the single-answer setting is shown in Figure \ref{fig:quality_results}. As shown in the figure, our method does not degrade the model's general language ability but even improves it in most categories. Note that the performance of both baselines on Coding and Math questions is poor because the AlpacaGPT-4 dataset does not have a strong specialty in questions of this category. When evaluating under the pairwise setting, MCTune achieves a 51.25\% win rate over instruction fine-tuning. These results suggest that training for controllability is beneficial for improving LLMs' natural language performance.

\subsection{Analyses}
\subsubsection{Relationship Between Linguistic Controllability and Generation Quality}
This experiment focuses on studying the relationship between generation quality and the controllability error of LLM responses. To start off, we randomly select a set of 5 linguistic controls $C$ and fix it throughout the experiment for simplicity. Given an LLM $p_\theta$ and the Alpaca-GPT4 test dataset $D_\text{test} = \left\{(\mathbf{x}_i, \mathbf{y}_i)\right\}^{N}_{i=1}$, we select 50 examples from $D_\text{test}$ then sample 50 responses $\hat{\mathbf{y}}_i$ from $p_\theta(\mathbf{y}_i\mid \mathbf{x}_i, f_C(\mathbf{y}_i))$. Each $\hat{\mathbf{y}}_i$ is then evaluated by GPT-4 Turbo on how well it satisfies $\mathbf{x}_i$ and a controllability error of $\hat{\mathbf{y}}_i$ is computed by taking the average of the normalized $L_1$ errors across all controls in $C$. We also consider three different settings where $\sigma=0.1$, $\sigma=0.3$, and $\sigma=0.5$ to see how the pattern shifts as $\sigma$ changes. The results are shown in Figure \ref{fig:quality_vs_controllability}. We can see that, compared to IFT, our method achieves better quality and controllability simultaneously. As $\sigma$ increases, the controllability error increases, which is expected and consistent with the results in Section \ref{fig:ablation_sigma}. We then start to notice a slight degradation in quality at $\sigma=0.5$, where there are no responses with a score of 10, and more responses with low scores begin to appear. This observation suggests that there might be a positive correlation between controllability and generation quality.

\subsubsection{Analysis of Model Controllability with Varying Linguistic Controls} \label{sec:ablation_num_tags}
\begin{figure}[h]
\begin{center}
\centerline{\includegraphics[width=\columnwidth]{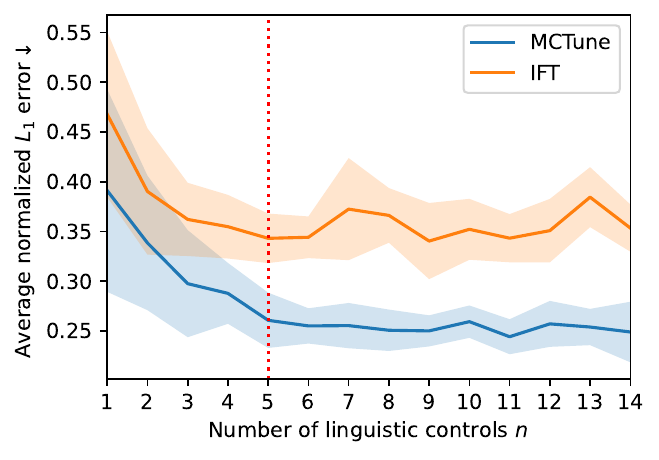}}
\caption{Linguistic controllability error on the test set as the number of linguistic controls $n$ per sample in MCTune's training increases. The solid curves represent the linguistic controllability error averaged over all linguistic complexities (lower is better) with the shaded areas represent the 95\%-confidence interval. The dotted vertical line indicates the maximum number of controls (5) used during MCTune's training.}
\label{fig:ablation_num_tags}
\end{center}
\end{figure}
In this experiment, we are interested in studying how the model's controllability is affected as the number of linguistic controls increases. We conduct this experiment using the Alpaca-GPT4 dataset, from which we randomly sample 100 examples from the test set. For each example, we sample five new control vectors $f'$ of length $n$, compute the average normalized $L_1$ error per control, and then average these over all controls to obtain a single numeric score. We repeat this process for $n=1$ to $n=14$. The results are shown in Figure \ref{fig:ablation_num_tags}. Surprisingly, controllability is worse when $n$ is small. This is true for both MCTune and the IFT baseline, suggesting that this is not an inherent limitation of our approach. Notice that, as the number of linguistic controls increases beyond five, the maximum number of controls used during training, the controllability error of IFT fluctuates, while MCTune remains approximately stable. This demonstrates that our model, trained with MCTune, is able to generalize to a larger number of controls than it has seen in training.

\subsubsection{Analyzing the Impact of $\sigma$ on Linguistic Controllability Evaluation} \label{sec:sigma_impact}
\begin{figure}[h]
\begin{center}
\centerline{\includegraphics[width=6cm]{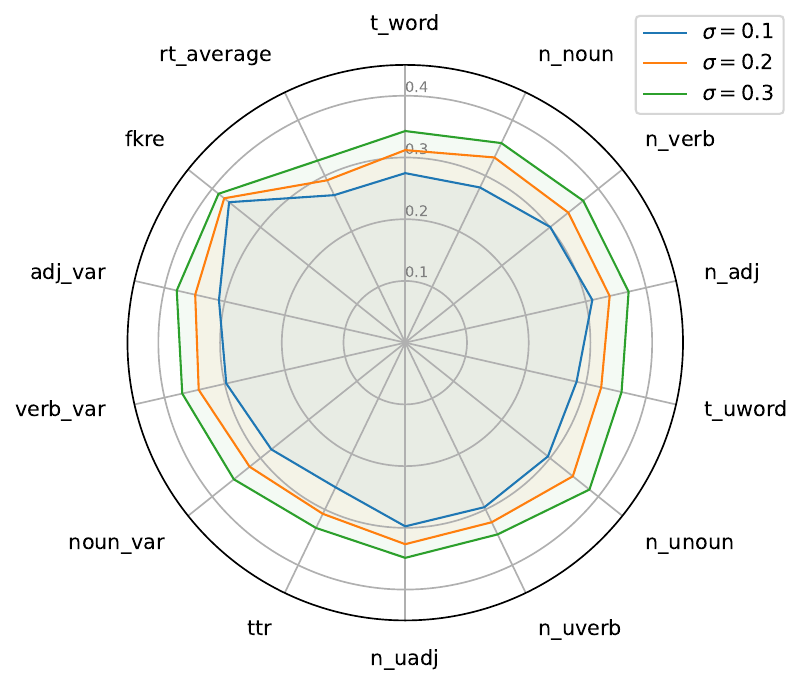}}
\caption{Analysis of how the average normalized $L_1$ error varies when \textbf{changing the test difficulty level} via $\sigma$. We use the same model and only vary $\sigma$.}
\label{fig:ablation_sigma}
\end{center}
\end{figure}
This section examines the effect of $\sigma$ on controllability evaluation. Given a model fine-tuned with MCTune and a test dataset, we follow the process described in Section \ref{control_eval} to sample new linguistic control vectors for evaluating controllability. We consider three settings: $\sigma=0.1$, $\sigma=0.2$, and $\sigma=0.3$. The results are depicted in Figure \ref{fig:ablation_sigma}. As illustrated by the results, as $\sigma$ increases, the average normalized $L_1$ error uniformly increases across all linguistic controls. This indicates that $\sigma$ can serve as a hyperparameter to adjust the difficulty of the evaluation, which is intuitive since a larger $\sigma$ enables the sampling of control vectors that are more distant from the reference control vector $f(\mathbf{y})$.

\subsection{Case Studies}
\begin{figure}[h]
\begin{center}
\centerline{\includegraphics[width=0.97\columnwidth]{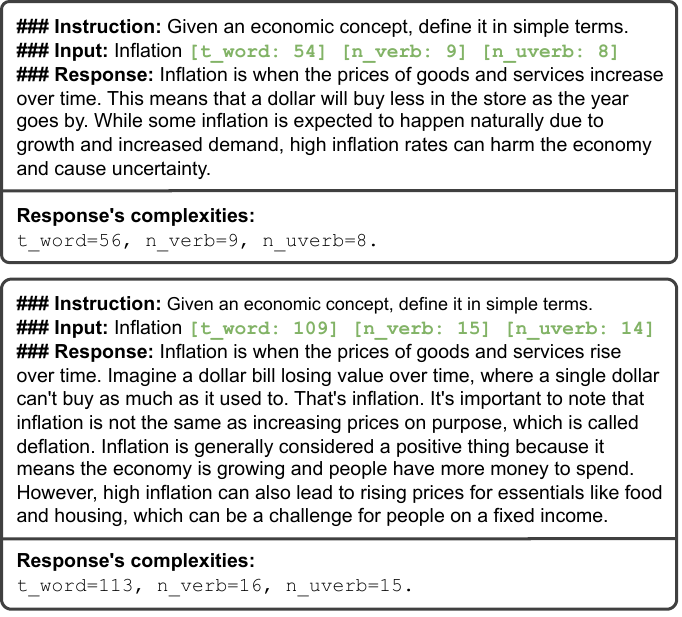}}
\caption{Examples of MCTune model responding to the same instruction under different linguistic controls. The linguistic complexities of the responses are extracted using LFTK with exact values shown at the bottom of each example.\looseness-1}
\label{fig:case_study}
\end{center}
\end{figure}
This section presents examples of how different values of controls affect the model's responses. As shown in Figure \ref{fig:case_study}, increasing the \texttt{t\_word} control leads to a longer response and larger values of \texttt{n\_verb} and \texttt{n\_uverb} increase the variation in verb choice. Although it is challenging to strictly control the complexities of the response, the linguistic controls can serve as soft constraints on the model's generation.

\section{Discussion and Conclusions}
In this paper, we advance the precise control of linguistic complexities in LLMs through multi-objective control tuning. Our method is straightforward and can be seamlessly integrated with existing instruction-tuning datasets without the need to gather new ones. Our training objective concurrently optimizes the LLM's ability to follow instructions and its controllability. Through our experiments, we find that incorporating this dual-focus strategy significantly improves the LLM's generative quality, surpassing the results of instruction fine-tuning alone. This finding suggests that controllability and instruction-following ability may have a complementary effect on each other. Additionally, we observe that while state-of-the-art LLMs achieve impressive natural language performance, they are not easy to control. This emphasizes the need for studying methods that improve controllability in LLMs.

\section{Limitations}
While our method significantly improves controllability compared to regular instruction fine-tuning, we observe that there is still room for improvement. The model trained with MCTune is able to loosely follow the linguistic controls but struggles to produce responses with the exact complexities. An interesting next step would be to improve controllability in a strict setting. Another limitation we observe is the fact that LFTK sometimes extracts incorrect linguistic complexities for a given text. This leads to noisy controls that may confuse and reduce the controllability of the model during training.\looseness-1

\bibliography{acl}
\bibliographystyle{acl_natbib}

\appendix

\section{Computing Linguistic Features} \label{sec:compute_features}
In this section, we describe in greater detail how linguistic features are computed. We use LFTK as a feature extractor, which is based on spaCy. Given a string, spaCy tokenizes it into a sequence of tokens along with their annotations (e.g., part-of-speech). The number of words \texttt{t\_word} is the number of tokens. The \texttt{n\_noun}, \texttt{n\_verb}, and \texttt{n\_adj} are computed based on the POS provided by spaCy. The \texttt{t\_uword}, \texttt{n\_unoun}, \texttt{n\_uverb}, and \texttt{n\_uadj} are computed accordingly. The type-token ratio, noun variation, verb variation, and adjective variation are $\frac{\texttt{t\_uword}}{\texttt{t\_word}}$, $\frac{\texttt{n\_unoun}}{\texttt{n\_noun}}$, $\frac{\texttt{n\_uverb}}{\texttt{n\_verb}}$, and $\frac{\texttt{n\_uadj}}{\texttt{n\_adj}}$, respectively. Let \texttt{t\_sent} be the number of sentences and \texttt{t\_syll} is the number of syllables, the formula to compute Flesch-Kincaid reading ease is
$$206.835 - 1.015\left(\frac{\texttt{t\_word}}{\texttt{t\_sent}}\right) - 84.6\left(\frac{\texttt{t\_syll}}{\texttt{t\_word}}\right)$$
Lastly, the average reading time \texttt{rt\_average} is $\frac{\texttt{t\_word}}{240}$.

\section{Verifying Validity of Control Vectors} \label{sec:verify_validity}
Given a linguistic control vector, it is straightforward to check its validity by iterating through a list of pre-defined rules.
\begin{itemize}
    \item $\texttt{t\_word} > 0$
    \item $\texttt{t\_word} \geq \texttt{n\_noun}+\texttt{n\_verb}+\texttt{n\_adj}$
    \item $\texttt{t\_word} \geq \texttt{t\_uword}$
    \item $\texttt{n\_noun} \geq 0$
    \item $\texttt{n\_noun} \geq \texttt{n\_unoun}$
    \item $\texttt{n\_verb} \geq 0$
    \item $\texttt{n\_verb} \geq \texttt{n\_uverb}$
    \item $\texttt{n\_adj} \geq 0$
    \item $\texttt{n\_adj} \geq \texttt{n\_uadj}$
    \item $\texttt{t\_uword} > 0$
    \item $\texttt{t\_uword} \geq \texttt{n\_unoun}+\texttt{n\_uverb}+\texttt{n\_uadj}$
    \item $\texttt{n\_unoun} \geq 0$
    \item $\texttt{n\_uverb} \geq 0$
    \item $\texttt{n\_uadj} \geq 0$
    \item $\texttt{fkre} \leq 121.22 $
\end{itemize}
If any of the rules above is violated, we conclude that the control vector is invalid.

\begin{table*}[]
\centering
\resizebox{\textwidth}{!}{%
\begin{tabular}{lrrrrrrrrrrrrrr}
\toprule
Method              & t\_word       & n\_noun       & n\_verb       & n\_adj        & t\_uword      & n\_unoun      & n\_uverb      & n\_uadj       & ttr           & noun\_var     & verb\_var     & adj\_var      & fkre          & rt\_average   \\
\midrule
IFT (Zero-shot)     & 0.31          & 0.33          & 0.34          & 0.32          & 0.38          & 0.35          & 0.33          & 0.35          & 0.30          & 0.30          & 0.31          & 0.32          & 0.34          & 0.31          \\
IFT (Few-shot)      & 0.33          & 0.32          & 0.33          & 0.26          & 0.39          & 0.41          & 0.31          & 0.28          & 0.38          & 0.30          & 0.33          & 0.32          & 0.39          & 0.31          \\
GPT-3.5 (Zero-shot) & 0.37          & 0.50          & 0.36          & 0.44          & 0.55          & 0.34          & 0.33          & 0.54          & 0.26          & 0.33          & 0.48          & 0.29          & 0.39          & 0.47          \\
GPT-3.5 (Few-shot)  & 0.48          & 0.53          & 0.65          & 0.37          & 0.67          & 0.60          & 0.39          & 0.44          & 0.40          & 0.29          & 0.35          & 0.28          & 0.40          & 0.48          \\
GPT-4 (Zero-shot)   & 0.64          & 0.52          & 0.34          & 0.51          & 0.63          & 0.73          & 0.46          & 0.39          & \textbf{0.21} & 0.32          & 0.38          & 0.28          & 0.49          & 0.51          \\
GPT-4 (Few-shot)    & 0.40          & 0.40          & 0.47          & 0.31          & 0.46          & 0.51          & 0.35          & 0.36          & 0.36          & 0.28          & 0.30          & 0.27          & 0.40          & 0.32          \\
RLHF (REINFORCE)    & 0.25          & 0.24          & 0.27          & 0.40          & 0.27          & 0.25          & 0.25          & 0.27          & 0.29          & 0.29          & 0.38          & 0.30          & 0.83          & 0.36          \\
MCTune              & \textbf{0.18} & \textbf{0.22} & \textbf{0.25} & \textbf{0.25} & \textbf{0.21} & \textbf{0.23} & \textbf{0.22} & \textbf{0.25} & 0.27          & \textbf{0.26} & \textbf{0.29} & \textbf{0.26} & \textbf{0.31} & \textbf{0.20} \\
\bottomrule
\end{tabular}%
}
\caption{Additional comparison of controllability error with GPT-4, few-shot and RL-based baselines.}
\label{tab:additional_exp1}
\end{table*}
\begin{table*}[h]
\centering
\resizebox{\textwidth}{!}{%
\begin{tabular}{lrrrrrrrrrrrrrr}
\toprule
Method & t\_word       & n\_noun       & n\_verb       & n\_adj        & t\_uword      & n\_unoun      & n\_uverb      & n\_uadj       & ttr           & noun\_var     & verb\_var     & adj\_var      & fkre          & rt\_average   \\
\midrule
IFT    & 0.37          & 0.37          & 0.33          & 0.35          & 0.41          & 0.38          & 0.33          & 0.35          & \textbf{0.44} & \textbf{0.40} & 0.43          & 0.47          & 0.40          & 0.36          \\
MCTune & \textbf{0.34} & \textbf{0.32} & \textbf{0.32} & \textbf{0.29} & \textbf{0.37} & \textbf{0.35} & \textbf{0.32} & \textbf{0.34} & 0.47          & 0.41          & \textbf{0.42} & \textbf{0.44} & \textbf{0.33} & \textbf{0.31} \\
\bottomrule
\end{tabular}%
}
\caption{Comparison of controllability error when training Mistral-7B-v0.1 with IFT and MCTune (ours).}
\label{tab:additional_exp2}
\end{table*}

\section{Additional Experiments With More Baselines} \label{sec: additional_baselines}

In this section, we conduct experiments with other baselines that include:

\paragraph{GPT-4.} Similar to other baselines, we query GPT-4 (\texttt{gpt-4-0125-preview}) using the prompt template in Figure \ref{fig:prompt_template} in a zero-shot manner. Since GPT-4 is very expensive, we only evaluate it on 500 test examples.

\paragraph{Few-shot baselines.} To introduce stronger baselines, we consider the few-shot setting where a few exemplars are included in the prompt to boost the performance of the underlying model. We carefully select 5 exemplars from the training set that can cover all the linguistic complexities.

\paragraph{RL-based methods.} We compare with a variant of RLHF proposed in \citet{DBLP:journals/corr/abs-2402-14740}, where instead of training the LLM policy with PPO, they use REINFORCE. To adapt this method to our task, we use the negative controllability error as the rewards. Since they can be computed deterministically, we do not need to use neural networks for reward modeling. More specifically, given $(\mathbf{x}, \mathbf{y}, f_C)\sim D$ and a response $\hat{\mathbf{y}}$, the reward is $$R_{f_C}(\hat{\mathbf{y}}, \mathbf{y}) = -\frac{1}{|C|}\sum_{i=1}^{|C|}\frac{|f_{C_i}(\hat{\mathbf{y}})-f_{C_i}(\mathbf{y})|}{M_i - m_i}$$ where $M_i$ and $m_i$ are the maximum and minimum values of the $i$-th linguistic complexity in the training set. The REINFORCE training objective is: $$\mathbb{E}_{(\mathbf{x}, \mathbf{y}, f_C)\sim D}\left [ (R_{f_C}(\hat{\mathbf{y}}, \mathbf{y})-b)\nabla_{\theta}\log{p_{\theta}(\hat{\mathbf{y}}|\mathbf{x})} \right ]$$ where $\hat{\mathbf{y}}\sim p_{\theta}(\cdot |\mathbf{x}, f_C(\mathbf{y}))$ and $b$ is the average of all achieved rewards in previous steps. For the Fine-grained RLHF, we are actively adapting their codebase to our task and will report the results shortly. As shown in Table \ref{tab:additional_exp1}, MCTune exhibits the lowest controllability errors on most linguistic complexities.

\section{Additional Experiments With Other LLM Architectures} \label{sec:additional_architectures}

In this section, we evaluate our method on Mistral-7B-v0.1 \cite{DBLP:journals/corr/abs-2310-06825} to see how it generalizes to LLM architectures other than LLaMA. As shown in Table \ref{tab:additional_exp2}, MCTune keeps improving the controllability of Mistral on most attributes.

\end{document}